\documentclass[10pt,twocolumn,letterpaper]{article}

\usepackage{wacv}
\usepackage{times}
\usepackage{epsfig}
\usepackage{graphicx}
\usepackage{amsmath}
\usepackage{amssymb}
\usepackage{booktabs}
\usepackage{color}
\newcommand*{\affmark}[1][*]{\textsuperscript{#1}}
\usepackage[subrefformat=parens]{subcaption}
\usepackage[accsupp]{axessibility}

\def\wacvPaperID{1623} 

\wacvfinalcopy 

\ifwacvfinal
\usepackage[breaklinks=true,bookmarks=false]{hyperref}
\else
\usepackage[pagebackref=true,breaklinks=true,colorlinks,bookmarks=false]{hyperref}
\fi

\begin{document}

\title{Rethinking Rotation in Self-Supervised Contrastive Learning: \\Adaptive Positive or Negative Data Augmentation}


\author{Atsuyuki Miyai\affmark[1]\ \ \ Qing Yu\affmark[1]\ \ \ Daiki Ikami\affmark[2]\ \ \ Go Irie\affmark[2]  \ \ Kiyoharu Aizawa\affmark[1] \\
 \affmark[1]The University of Tokyo\ \ \ \ \affmark[2]NTT Corporation, Japan\\ 
{\tt\small \{miyai,yu,aizawa\}@hal.t.u-tokyo.ac.jp}\ \ \ \ 
\tt\small {daiki-ikami@go.tuat.ac.jp}\ \ \ \  \tt\small{goirie@ieee.org}
}



\maketitle
\thispagestyle{empty}

\begin{abstract}
   Rotation is frequently listed as a candidate for data augmentation in contrastive learning but seldom provides satisfactory improvements. We argue that this is because the rotated image is always treated as either positive or negative. The semantics of an image can be rotation-invariant or rotation-variant, so whether the rotated image is treated as positive or negative should be determined based on the content of the image. Therefore, we propose a novel augmentation strategy, adaptive Positive or Negative Data Augmentation (PNDA), 
   in which an original and its rotated image are a positive pair if they are semantically close and a negative pair if they are semantically different. To achieve PNDA, we first determine whether rotation is positive or negative on an image-by-image basis in an unsupervised way. Then, we apply PNDA to contrastive learning frameworks. Our experiments showed that PNDA improves the performance of contrastive learning. The code is available at \url{ https://github.com/AtsuMiyai/rethinking_rotation}.
\end{abstract}

\section{Introduction}
\begin{figure}[t]
  \centering
  \includegraphics[keepaspectratio, scale=0.31]
  {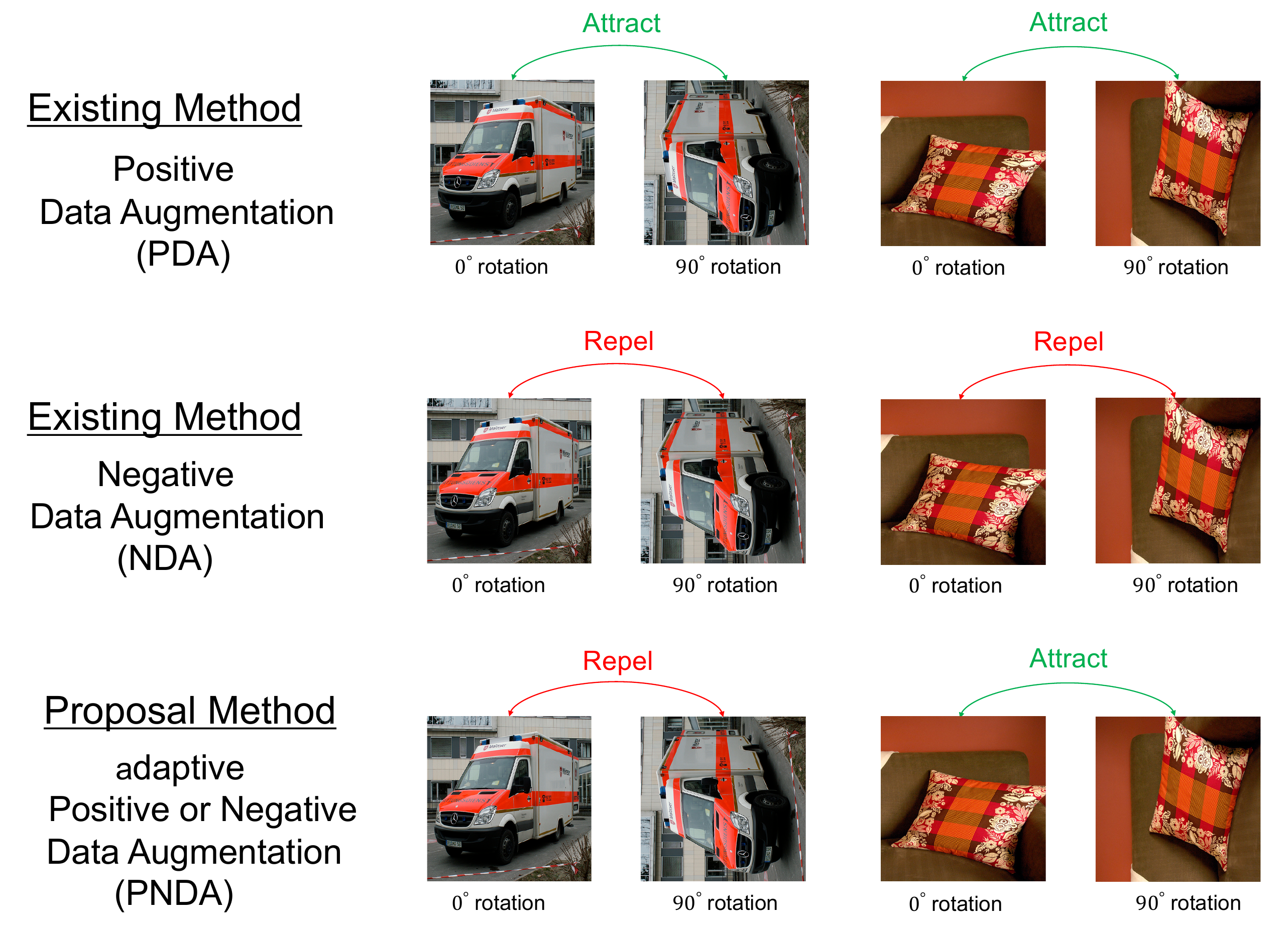}
  \caption{\textbf{Comparison of previous and the
proposed augmentation strategy.} \textbf{Upper:} PDA treats all rotated images as positives and encourages them to be pulled closer. \textbf{middle:} NDA treats all rotated images as negatives and encourages them to be pushed away. \textbf{Lower:} Our proposed PNDA considers the semantics of the images, and treats rotation as either positive or negative for each image.}
  \label{fig:proposal_augmentation}
\end{figure}

Recently, self-supervised learning~\cite{noroozi_unsupervised_2016, gidaris2018unsupervised, He_2020_CVPR, chen2020simple, he2021masked} has shown remarkable results in representation learning. The gap between self-supervised and supervised learning has been 
bridged by contrastive learning~\cite{He_2020_CVPR, chen2020simple, grill2020bootstrap, chen2020simsiam, caron2020unsupervised, caron2021emerging}. For self-supervised contrastive learning, data augmentation is one of the most important techniques~\cite{tian2020makes}.
A common approach for contrastive learning creates positives with some augmentations and encourages them to be pulled closer. Since this augmentation strategy creates positive samples, we refer to it as positive data augmentation (PDA). 
In addition, some methods~\cite{sinha2021negative, tack2020csi, ge2021robust} use augmentation to create negatives and encourage them to be pushed away. This augmentation strategy is called negative data augmentation (NDA). 

Rotation has been attempted to be used for these augmentations, but few improvements have been made. Although rotation is useful in various fields, Chen \textit{et al.} \cite{chen2020simple} reported that rotation PDA degrades the representation ability in self-supervised contrastive learning because rotation largely affects image semantics. Since then, rotation has been treated as harmful for self-supervised contrastive learning.
We consider that this is because previous approaches tried treating rotation as either positive or negative without considering the semantics of each image.

To solve this problem and make full use of rotation, it is important to consider whether the rotation affects the semantics of each image.
Natural images are divided into two classes: rotation-agnostic image (RAI) with an ambiguous orientation and non-rotation-agnostic image (non-RAI) with a clear orientation. 
In RAI, an object can have various orientations. By applying rotation PDA to RAI and encouraging them to be pulled closer, the image will obtain embedding features robust to rotation. 
On the other hand, in non-RAI, the orientation of an object is limited. By applying rotation PDA to non-RAI and encouraging them to be pulled closer, the images will lose orientation information and might get undesirable features. For non-RAI, it is preferable to treat rotation as negative to maintain orientation information.

Based on this observation, in this study, we introduce a novel augmentation strategy called adaptive Positive or Negative Data Augmentation (PNDA). In Fig.\ref{fig:proposal_augmentation}, we show an overview of PDA, NDA, and PNDA. While PDA and NDA do not consider the semantics of each image, our proposed PNDA considers the semantics of each image, and treats rotation as positive if the original and rotated images have the same semantics and negative if their semantics are different. 
To achieve PNDA, we extract RAI for which rotation is treated as positive. However, there is no method to determine whether an image is RAI or non-RAI. Thus, we also tackle a novel task for sampling RAI and propose an entropy-based method. This sampling method focuses on the difference in the difficulty of the rotation prediction between RAI and non-RAI and can extract RAI based on the entropy of the rotation predictor’s output. 

We evaluate rotation PNDA with contrastive learning frameworks such as MoCo v2 and SimCLR. As a result of several experiments, we showed that the proposed rotation PNDA improves the performance of contrastive learning, while rotation PDA and NDA might decrease it.  

The contributions of our paper are summarized as follows:
\begin{itemize}
      \item We propose a novel augmentation strategy called PNDA that considers the semantics of the images and treats rotation as the better one of either positive or negative for each image.
      \item We propose a new task of sampling rotation-agnostic images for which rotation is treated as positive.
      \item We apply rotation PNDA with contrastive learning frameworks, and found that rotation PNDA improves the performance of contrastive learning.
\end{itemize}

\section{Related work}
\label{sec:related_work}
\subsection{Contrastive Learning}
Contrastive learning has become one of the most successful methods in self-supervised learning~\cite{He_2020_CVPR, chen2020simple, grill2020bootstrap, chen2020simsiam, caron2021emerging}.
One popular approach for contrastive learning, such as MoCo~\cite{He_2020_CVPR} and SimCLR~\cite{chen2020simple}, is to create two views of the same image and attract them while repulsing different images. Many studies have explored the positives or negatives of MoCo and SimCLR~\cite{Dwibedi_2021_ICCV, Zheng_2021_ICCV, huynh20a}. Some methods, such as BYOL~\cite{grill2020bootstrap} or SimSiam~\cite{chen2020simsiam}, use only positives, but recent studies~\cite{ge2021robust, Wang_2021_ICCV} have shown that better representation can be learned by incorporating negatives into these methods. For contrast learning, the use of positives and negatives is important to learn better representations.

\subsection{Data Augmentation for Contrastive Learning}
There are two main types of augmentation strategies for contrastive learning: positive data augmentation (PDA) and negative data augmentation (NDA). 
\subsubsection{Positive Data Augmentation (PDA)}
Contrastive learning methods create positives with augmentations and get them closer. For example, Chen \textit{et al.}~\cite{chen2020simple} proposed composition of data augmentations \textit{e.g.} Grayscale, Random Resized Cropping, Color Jittering, and Gaussian Blur to make the model robust
to these augmentations. On the other hand, they reported that adding rotation to these augmentations degrades performance. However, they used rotation PDA without considering the difference in the semantic content between RAI and non-RAI. Some work dealt with rotation for contrastive learning by residual relaxation~\cite{wang2021residual} or combination with rotation prediction~\cite{addepalli2022towards, dangovski2021equivariant}. Our work focuses on the semantics of each rotated image.
\subsubsection{Negative Data Augmentation (NDA)}
Several methods have been proposed to create negative samples by applying specific transformations to images~\cite{ijcai2021-84, tack2020csi, sinha2021negative}. Sinha \textit{et al.}~\cite{sinha2021negative} investigated whether several augmentations, including Cutmix~\cite{Yun2019CutMixRS} and Mixup~\cite{zhang2018mixup}, which are typically used as positive in supervised learning, can be used as NDA for representation learning. However, they did not argue that rotation NDA is effective. Tack \textit{et al.}~\cite{tack2020csi} stated rotation NDA is effective for unsupervised out-of-distribution detection, but they also did not state that rotation NDA is effective for representation learning. These methods~\cite{ijcai2021-84, tack2020csi, sinha2021negative} treat the transformed images as negatives without considering the semantics of each image.

\subsection{Rotation Invariance}
Rotation invariance is one of many good and well-studied properties of visual representation, and many existing methods incorporate rotational invariant features into feature learning frameworks. For supervised learning, G-CNNs~\cite{cohen2016group} and Warped Convolutions~\cite{henriques2017warped} showed excellent results in learning rotational invariant features. For self-supervised learning, 
Feng \textit{et al.}~\cite{Feng_2019_CVPR} worked on rotation feature learning, which learns a representation that decouples rotation related and unrelated parts. However, previous works separated the rotation related and unrelated parts implicitly as internal information of the network and did not explicitly extract RAI. Here, we tackle a novel task for sampling RAI.

\vskip\baselineskip
In this paper, we propose a novel augmentation strategy called PNDA that considers the semantics of the image and treats rotation as positive for RAI and negative for non-RAI. To achieve PNDA, we also tackle a novel task for sampling RAI. We demonstrated the effectiveness of rotation PNDA with contrastive learning frameworks with sampled RAI and non-RAI.

\section{Rotation-agnostic Image Sampling}

\begin{figure*}[t]
  \centering
  \includegraphics[keepaspectratio, scale=0.14]
  {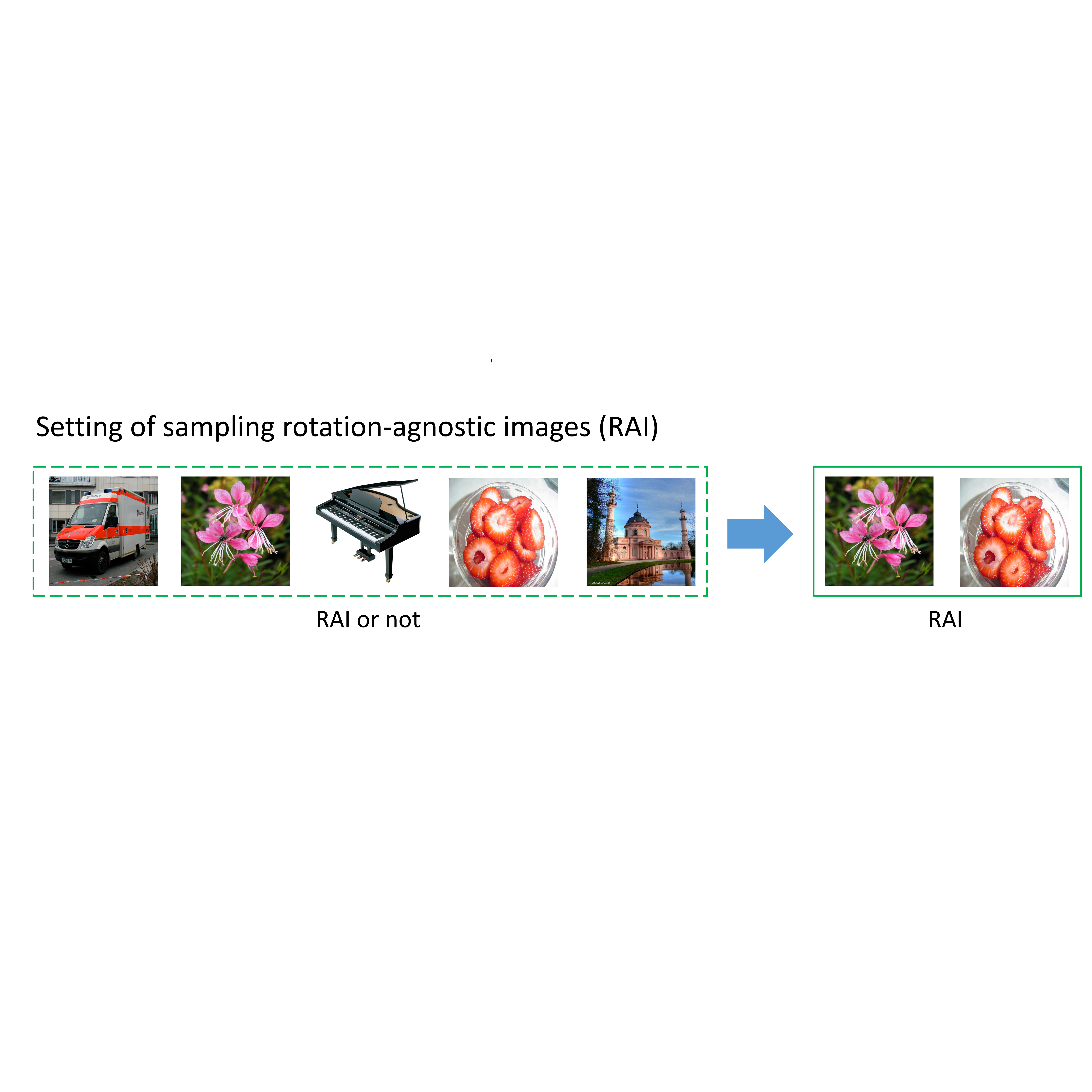}
  \caption{\textbf{The setting of RAI sampling}. We have the set of RAI and non-RAI images. Our goal is to extract RAI in an unsupervised way.}
  \label{fig:setting_sampling}
\end{figure*}

\label{sec:rais}
\begin{figure*}[t]
  \centering
  \includegraphics[keepaspectratio, scale=0.48]
   {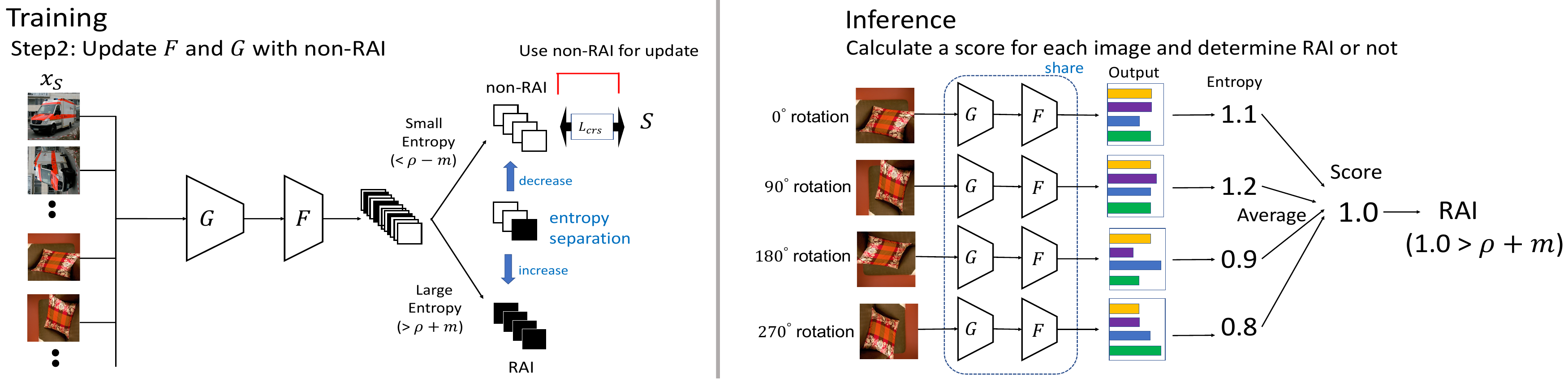}
   \caption{\textbf{Overview of training steps and inference step in the proposed sampling method.} $G$ is a feature extractor and $F$ is a rotation predictor. During training, at Step1, we initialize the network with all samples before overfitting. At Step2, we update the network using only non-RAI and make a boundary between non-RAI and RAI. At inference, we calculate the score by averaging the entropy of $F$'s outputs of 4 rotated images, and we determine RAI or not.}
   \label{fig:proposal_rais}
 \end{figure*}
To achieve PNDA, we first need to extract RAI for which rotation is treated as positive.
In Fig.~\ref{fig:setting_sampling}, we show the setting of extracting RAI.
We have data of RAI and non-RAI, and our goal here is to extract RAI in an unsupervised way. In this section, we present our novel entropy-based method for sampling RAI. First, we illustrate the overall concept of the method in Section~\ref{subsec:overall_concept}. Then, in Section~\ref{subsec:training_procedure}, we detail the training procedure, and in Section~\ref{subsec:inference} we explain the inference procedure. Finally, we explain the criterion of tuning the hyperparameters in Section~\ref{subsec:criterion_for_tune}.
\subsection{Overall Concept}
\label{subsec:overall_concept}
This sampling method focuses on the difference in the difficulty of the rotation prediction between RAI and non-RAI. For RAI, the feature distributions of the original and rotated images are similar, so the model can hardly predict which rotation is applied. Thus, the entropy of the rotation predictor's outputs should be large for RAI. On the other hand, for non-RAI, the feature distributions of the original and rotated images are different, so the model can easily predict which rotation is applied. Hence, the entropy of the rotation predictor's outputs should be small for non-RAI. Therefore we can separate RAI and non-RAI by the entropy of the rotation predictor.

We show an overview of our approach in Fig.\ref{fig:proposal_rais}. $G$ is a feature generator network, and $F$ is a rotation predictor network. Our idea is to train the rotation predictor $F$ to learn the boundary between RAI and non-RAI. The key is to update the rotation predictor $F$ using only non-RAI and to create a rotation predictor that can correctly predict the rotation of only non-RAI.

\subsection{Training Procedure}
\label{subsec:training_procedure}
From the previous discussion in Section~\ref{subsec:overall_concept}, we propose a training procedure consisting of the following two steps, as shown in Fig.~\ref{fig:proposal_rais}. 
\subsubsection{Step1.} At the first step, we train an initial model with all samples before overfitting. We define the set of transformations $\mathcal{S}$ as all the image rotations by multiples of 90 degrees, i.e., image rotations by 0, 90, 180, and 270 degrees. Namely, we denote $\mathcal{S} := \{S_0, S_{90}, S_{180}, S_{270}\}$. We apply $\mathcal{S}$ to a set of all images. We train the model to predict which transformation $S\in \mathcal{S}$ is applied. As preprocessing, for a given batch of samples $\mathcal{B} = \{x_i\}_{i=1}^B$, we apply $\mathcal{S}$ to $\mathcal{B}$. The objective function in this step is as follows.
\begin{equation}
  \mathcal{L}_{\rm{crs}} =  \frac{1}{B}{\sum_{S\in \mathcal{S}}}\, {\sum_{x_{S}\in \mathcal{B}_S}}-S\log\left(p\left(x_S\right)\right), \;
      \mathcal{B}_S = \{S(x_i)\}_{i=1}^B.
      \label{eq:rotation_pred}
\end{equation}
$p\left(x_S\right)$ denotes the $|\mathcal{S}|$-dimensional softmax class probabilities for input $x_S$.
We train the model at this step for $\beta_1$ epochs.
\subsubsection{Step2.}
We propose a separation loss to separate RAI from non-RAI. Specifically, we first define the following two losses.
\footnotesize
\begin{equation}
\begin{aligned}  
      \mathcal{L}_{\rm{es}} &= \frac{1}{B}{\sum_{S\in \mathcal{S}}}\, {\sum_{x_{S}\in \mathcal{B}_S}} \mathcal{L}_{\rm{es}}\left(p\left(x_S\right)\right), 
      \\
      \mathcal{L}_{\rm{es}}\left(p\right) &=
      \left\{
        \begin{array}{ll}
        -|H(p) - \rho|   & (|H(p) - \rho| > m), \\ 
          0  &\text{otherwise}.\\ 
        \end{array}
      \right.
      \label{eq:entropy_separtation}
\end{aligned}
\end{equation}
\normalsize

\footnotesize
\begin{equation}
\begin{aligned} 
\tilde{\mathcal{L}}_{\rm{crs}} &= \frac{1}{B}{\sum_{S\in \mathcal{S}}}\, {\sum_{x_{S}\in \mathcal{B}_S}} \tilde{\mathcal{L}}_{\rm{crs}}\left(p\left(x_S\right)\right),
\\
\tilde{\mathcal{L}}_{\rm{crs}}\left(p\right) &= 
      \left\{
        \begin{array}{ll}
          \mathcal{L}_{\rm{crs}}\left(p\left(x_S\right)\right) & ( H(p) - \rho < -m), \\ 
          0  &\text{otherwise}.\\ 
        \end{array}
        \right.
        \label{eq:rotation_clean}
\end{aligned}
\end{equation}
\normalsize

 Eq.(\ref{eq:entropy_separtation}) is the entropy separation loss proposed by \cite{saito2020dance}. $H(p)$ is the entropy of $p$. $\rho$ is set to $\frac{\log(|\mathcal{S}|)}{2}$, since $\log(|\mathcal{S}|)$ is the maximum value of $H(p)$. $m$ is the margin for separation. This loss enables the entropy of RAI to be larger and urges the rotation predictor to misclassify the rotation of RAI, and enables the entropy of non-RAI to be smaller, and urges the rotation predictor to predict the rotation of non-RAI more confidently. The loss in Eq.(\ref{eq:rotation_clean}) enables the model to learn using only non-RAI. With a hyperparameter $\lambda$, the final objective is as follows.
\begin{equation}
      \tilde{\mathcal{L}}_{\rm{crs}} + \lambda\mathcal{L}_{\rm{es}}.
      \label{eq:rotation_separtation}
\end{equation}
We train the model at this step for $\beta_2$ epochs.
$\lambda$ in Eq.(\ref{eq:rotation_separtation}) is proportional to the epoch number: $\lambda = \lambda' \frac{\rm{epoch}}{\beta_2}$, where $\lambda'$ is a constant number. 
\subsection{Inference}
\label{subsec:inference}
At inference, we take images with four different rotations as input and calculate the average entropy of the outputs as a score that represents the difficulty of rotation prediction. We treat the images whose score is larger than $\rho+m$ as RAI and the other images as non-RAI. We treat rotation as positive for RAI and as negative for non-RAI.

\subsection{Criterion for tuning hyperparameters}
\label{subsec:criterion_for_tune}
The way of tuning the hyperparameters $\lambda'$ and $m$ focuses on the rotation classification accuracy after Step1 and Step2. At Step1, we train the rotation predictor before overfitting. At Step2, we train the rotation predictor with non-RAI and separate the entropy of RAI and non-RAI largely and extract RAI. The rotation prediction accuracy of the rotation predictor after Step2 should be almost the same as after Step1 because the number of non-RAI and RAI does not change between Step1 and Step2. We tuned the hyperparameters $\lambda'$ and $m$ so that the accuracy of the rotation prediction after Step 2 is the same as that after Step 1.

\section{PNDA for contrastive learning}
\label{sec:pnda_contrastive_learning}
In this section, we explain how to apply PNDA to contrastive learning frameworks~\cite{chen2020simple, chen2020mocov2}. We first describe contrastive learning (i.e. the InfoNCE loss) in the context of instance discrimination. Next, we introduce our approach to applying PNDA to contrastive learning.
\subsection{Contrastive Learning}
\label{subsec:contrastive_lr}
InfoNCE loss (i.e. contrastive loss) is commonly used in instance discrimination problems~\cite{chen2020simple, He_2020_CVPR}. Given an encoder network $f$ and an image $x$, we denote the output of the network as $z = f(x)$. We use $z_i$ as the embedding of a sample $x_i$ and use $z_p$ as the embedding of its positive sample $x_p$. We use $z_n\in {N_i}$ as embeddings of negative samples. The InfoNCE loss is defined as follows: 
\footnotesize
\begin{equation}
  {\mathcal{L}_{i}}^{\rm{InfoNCE}}=  -\log \frac{\exp({z_i}^\top{z_p}/\tau)}{\exp({z_i}^\top{z_p}/\tau) + {\sum_{z_{n}\in N_i}}\exp({z_i}^\top{z_{n}}/\tau)},
      \label{eq:contrastive}
\end{equation}
\normalsize
where $\tau$ is a temperature parameter.

SimCLR~\cite{chen2020simple} and MoCo v2~\cite{chen2020mocov2} create two views of the same image $\hat{x_i}$, $\hat{{x_i}}^{+}$ with random augmentation aug($\cdot$). Formally, $\hat{x_i} = \mathrm{aug}(x_i)$ and $\hat{{x_i}}^{+} = \mathrm{aug}(x_i)$. These two views are fed through the encoder $f$ to obtain embeddings ${z_i} = f(\hat{x_i})$ and ${z_i}^{+} = f(\hat{{x_i}}^{+})$. They encourage ${z_i}$ and ${z_i}^{+}$ to be pulled closer. That is, when $\hat{x_i}$ is an anchor image, the positive sample is $\hat{{x_i}}^{+}$. 
For negative samples, MoCo v2 uses a large dictionary as a queue of negative samples. SimCLR randomly samples a mini-batch of $M$ examples and makes pairs of augmented examples $\hat{X}$ and $\hat{X}^{+}$. Formally, $\hat{X}= \{\hat{{x_i}}\}_{i=1}^M$ and $\hat{X}^{+} = \{\hat{{x_i}}^{+}\}_{i=1}^M$. The mini-batch size results in $2M$.
For negative samples, SimCLR uses the other $2(M - 1)$ augmented examples other than positives within the mini-batch.

\subsection{Contrastive Learning with PNDA}
\label{sec:contrastive_pnda}
Rotation PNDA treats rotated images as positives for RAI and negatives for non-RAI. 
We define the set of positive samples ${P_i}^r$ and negative samples ${N_i}^r$ which contain rotated images of an anchor $x_i$. To deal with multiple positive pairs, we use supervised contrastive loss~\cite{khosla2020supervised}. 
The extended InfoNCE loss for PNDA is defined as follows: 
\small
\begin{equation}
\begin{aligned}
  {\mathcal{L}_{i}}^{\rm{PNDA}}&=\\
  -\frac{1}{|{P_i}^r|}&\sum_{{z_p}\in {P_i}^r} \log \frac{\exp({z_i}^\top{z_p}/\tau)}{\sum_{{z'}\in {P_i}^r \cup {N_i}^r }\exp({z_i}^\top{z'}/\tau)}.
      \label{eq:pnda_for_contrastive}
\end{aligned}
\end{equation}
\normalsize

To give a more detailed explanation, we define Rot$(x, \theta)$ is an image that rotates image $x$ by $\theta$ degrees. The detail explanations about ${P_i}^r$ and ${N_i}^r$ for MoCo v2 and SimCLR are provided below:
\subsubsection{PNDA for MoCo v2.}
For MoCo v2, we refer to~\cite{ge2021robust}, which incorporates patch-based NDA into MoCo v2. We extended~\cite{ge2021robust} for rotation PNDA. The set of anchor images for PNDA is the same as vanilla MoCo v2. We set ${P_i}^r$ to $\hat{{x_i}}^{+}$, Rot($\hat{{x_i}}^{+}$, 90), Rot($\hat{{x_i}}^{+}$, 180) and Rot($\hat{{x_i}}^{+}$, 270) for RAI and $\hat{{x_i}}^{+}$ for non-RAI. We set ${N_i}^r$ to vanilla MoCo v2's negative samples for RAI and vanilla MoCo v2's negative samples, Rot($\hat{{x_i}}^{+}$, 90), Rot($\hat{{x_i}}^{+}$, 180) and Rot($\hat{{x_i}}^{+}$, 270) for non-RAI.


\subsubsection{PNDA for SimCLR.} 
To the best of our knowledge, there is no method that applies NDA to SimCLR for
representation learning. We prioritize batch-wise processing, which is the essential mechanism of SimCLR, and the ease of implementation.
Like vanilla SimCLR, we create $\hat{X}^{+}$ and $\hat{X}^{+}$. In addition, with $\theta_1$ and $\theta_2$ which are different degrees chosen randomly from \{90, 180, 270\}, we create two sets of rotated images $\hat{X}_{\theta_1}$ and $\hat{X}_{\theta_2}$. Formally, $\hat{X}_{\theta_1}= \{\mathrm{Rot}(\hat{{x_i}},{\theta_1} )\}_{i=1}^M$, $\hat{X}_{\theta_2}= \{\mathrm{Rot}(\hat{{x_i}}^{+},{\theta_2} )\}_{i=1}^M$. The mini-batch size results in $4M$. Like vanilla SimCLR, we use $\hat{X}$ and $\hat{X}^{+}$ as anchor images.
We set ${P_i}^r$ to $\hat{{x_i}}^{+}$, Rot($\hat{{x_i}}$, $\theta_1$) and Rot($\hat{{x_i}}^{+}$, $\theta_2$) for RAI and $\hat{{x_i}}^{+}$ for non-RAI. We set ${N_i}^r$ to the other $4(M - 1)$ augmented examples within the mini-batch for RAI and the other $4M - 2$ augmented examples, including rotated images of an anchor ${x_i}$, within the mini-batch for non-RAI. Note that, although the mini-batch size increases, the diversity of the images in the mini-batch does not change since we increase the data by rotation.

\section{Experiments}
\label{sec:exp}
\subsection{Datasets}
\label{subsec:datasets}
We use CIFAR-100~\cite{Krizhevsky09learningmultiple} and Tiny ImageNet~\cite{le2015tiny}, which are used in self-supervised setting~\cite{zheng2021ressl,ermolov2020whitening,wang2021residual, NS2021}. CIFAR-100 contains 50,000 training images and 10,000 test images scaled down to 32$\times$32 in 100 different classes. Tiny ImageNet contains 100,000 training images and 10,000 test images scaled down to 64$\times$64 in 200 different classes, which are drawn from the original 1,000 classes of ImageNet~\cite{deng2009imagenet}. 

\subsection{Rotation-agnostic Image Sampling}
\subsubsection{Implementation Details.}
\begin{figure}[t]
  \begin{minipage}[b]{0.45\linewidth}
    \centering
    \includegraphics[keepaspectratio, scale=0.32]{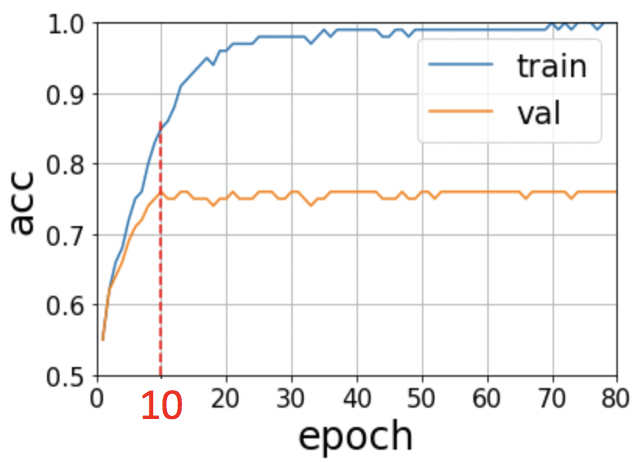}
    \subcaption{CIFAR-100}
  \end{minipage}
  \begin{minipage}[b]{0.55\linewidth}
    \centering
    \includegraphics[keepaspectratio, scale=0.32]{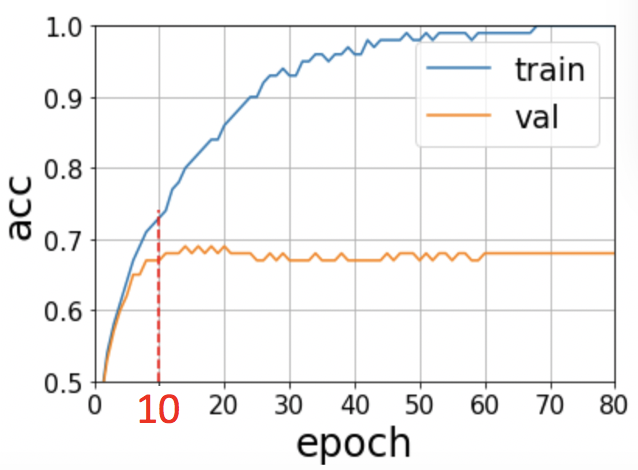}
    \subcaption{Tiny ImageNet}
  \end{minipage}
  \caption{The train and validation accuracy curves for rotation classification with split-training and split-validation data. We use the epoch just before overfitting to $\beta_1$.}
  \label{fig:train_test_acc}
\end{figure}

We used ResNet-18~\cite{7780459} as a feature encoder. Especially, for CIFAR-100, we use extended ResNet used by \cite{chen2020simple}. They replaced the first convolution layer with the convolutional layer with 64 output channels, the stride size of 1, the kernel size of 3, and the padding size of 3. They removed the first max-pooling from the encoder and added a non-linear projection head to the end of the encoder. In this study, we refer it to ResNet* to distinguish it from ResNet. 

To set $\beta_1$, we need to know the epoch before overfitting with all training data because self-supervised learning methods use all training data. However, to know the accurate epoch just before overfitting with all training data is impossible due to the lack of validation data. In order to know the approximate epoch just before overfitting, we treat 80\% of all training data as split-training data and treat the rest 20\% of all training data as split-validation data, and investigate the epoch just before overfitting with split-training data. We set $\beta_1$ to the epoch just before overfitting with split-training data, which is close to the epoch with all training data. In Fig.~\ref{fig:train_test_acc}, we show the train and validation accuracy curves with split-train and split-validation data for rotation classification. We set $\beta_1$ to 10 for both datasets.

We set $\beta_2$ to 200 for CIFAR-100 and 150 for Tiny ImageNet. According to Section~\ref{subsec:criterion_for_tune}, we set $\lambda'$ to 0.20 for CIFAR-100 and 0.10 for Tiny ImageNet and $m$ to 0.20 for both datasets. We use the Adam optimizer with a learning rate of 0.001 for CIFAR-100 and the Stochastic Gradient Descent (SGD) with a momentum of 0.9 and a learning rate of 0.1 for Tiny ImageNet. We use cosine decay schedule~\cite{loshchilov2017sgdr}. We train with a batch size of 64 in all experiments. We conducted 3 runs and chose the model that best matched the criterion in Section~\ref{subsec:criterion_for_tune}. We conducted the training on a single Nvidia V100 GPU. 
\subsubsection{Results on Rotation-agnostic Image Sampling.}
\label{subsubsec:results on_rais}
\begin{figure}[t]
  \centering
  \includegraphics[keepaspectratio, scale= 0.40]
  {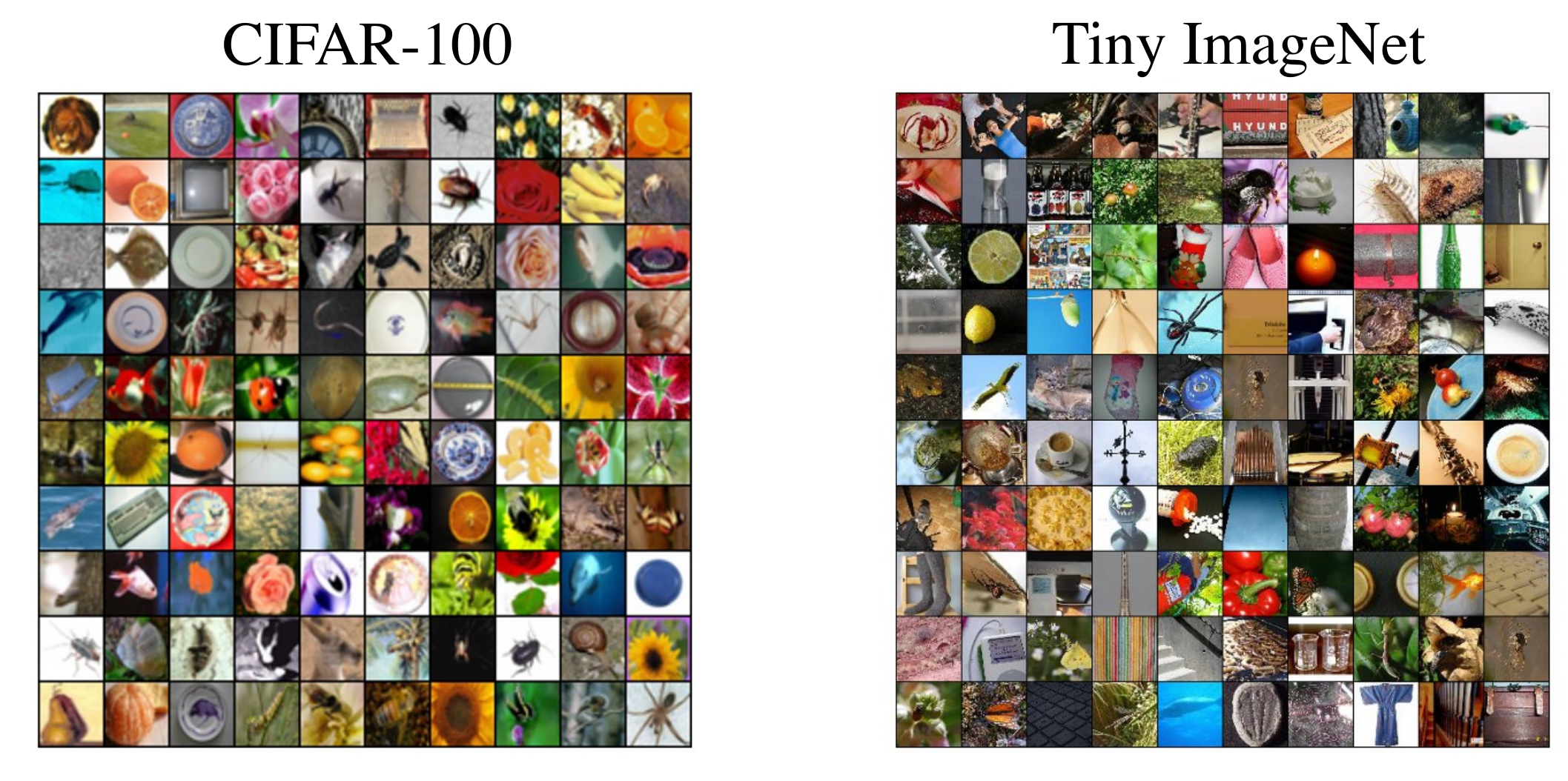}
  \caption{Examples of RAI extracted by our sampling framework on CIFAR-100 and Tiny ImageNet.}
  \label{fig:rais_visualize}
\end{figure}

\begin{table}[t]
  \centering
  {\tabcolsep = 1.0mm
  \caption{Number of RAI extracted by our sampling framework on CIFAR-100 and Tiny ImageNet}
  \scalebox{1.0}{
  \begin{tabular}{lccc}
  \\
    \hline
    Dataset   & \#RAI \; & \; \#images \; & ratio of RAI (\%)\\
    \hline \hline
    CIFAR-100 & \textbf{6,229} & 50,000 & \textbf{12.4}\\
    Tiny ImageNet~ & \textbf{30,711} & 100,000 & \textbf{30.7}\\
    \hline 
  \end{tabular}
  }
  \label{table:num_of_rais}
  }
\end{table}

\begin{figure}[t]
  \centering
  \includegraphics[keepaspectratio, scale= 0.38]
  {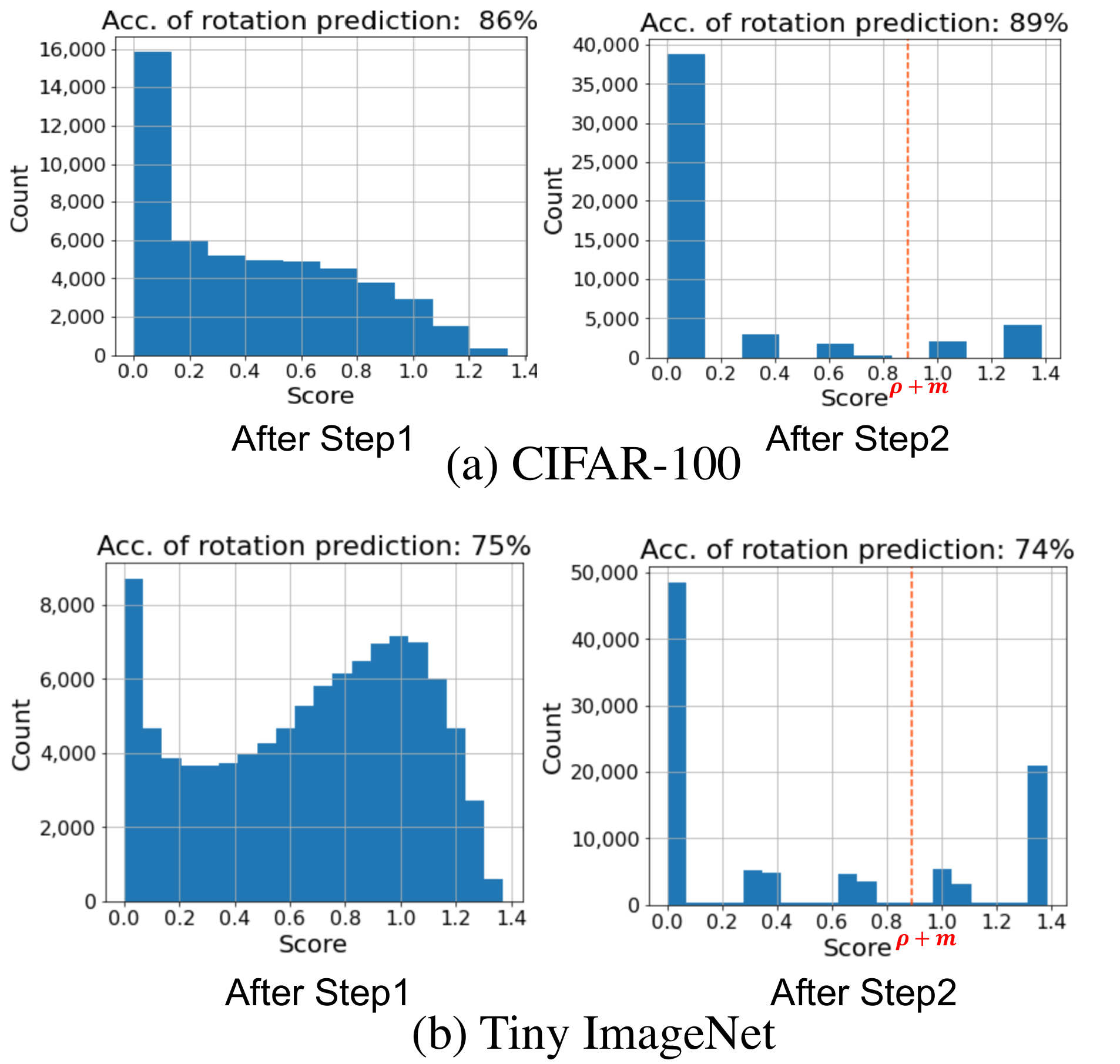}
  \caption{The histograms of the scores obtained with the model after Step1 and Step2 on CIFAR-100 and Tiny ImageNet. As Section~\ref{subsec:inference}, a score denotes the difficulty of predicting the rotation of an image. These results show that the model after Step2 ensures separation between non-RAI and RAI, whereas the model after Step1 confuses non-RAI and RAI.}
  \label{fig:comparison_entropy}
\end{figure}
  

In Fig.~\ref{fig:rais_visualize}, we show some examples of RAI extracted by our sampling framework on CIFAR-100 and Tiny ImageNet. This result shows that our sampling framework can extract RAI approximately correctly.
Table~\ref{table:num_of_rais} shows the number of RAI on CIFAR-100 and Tiny ImageNet. About 12\% for CIFAR-100 and about 31\% for Tiny ImageNet of all the images are extracted as RAI. Fig.\ref{fig:comparison_entropy} shows the histogram of the score of the outputs with the model after Step1 and Step2 on CIFAR-100 and Tiny ImageNet. As described in Section~\ref{subsec:inference}, a score denotes how difficult the rotation prediction is. Although the accuracies of rotation prediction with both models are almost the same, the distributions of scores are quite different. The model after Step2 makes the difference in the scores between RAI and non-RAI larger and enhances the separation, while the model after Step1 confuses non-RAI and RAI. 
Note that the number in Table~\ref{table:num_of_rais} is the number of RAI extracted by our sampling framework and not the actual number of RAI in the datasets. There are no ground truths of RAI and non-RAI, so the exact number of RAI is unknown.

\subsection{PNDA for contrastive learning}
\label{subsec:ex_pnda}
\subsubsection{Compared Methods.} We mainly use MoCo v2~\cite{chen2020mocov2} and SimCLR~\cite{chen2020simple} as contrastive learning frameworks. In addition to these baselines, we apply rotation PDA and rotation NDA to these frameworks. Rotation PDA regards all samples as RAI and treats rotated images as positives. Rotation NDA regards all samples as non-RAI and treats rotated images as negatives. 

\subsubsection{Evaluation Protocols.}
\label{subsec:eval_pro}
Following the previous works~\cite{He_2020_CVPR,chen2020simple}, we verify our methods by linear classification on frozen features, following a common protocol. After unsupervised pretraining, we freeze the features and train a supervised linear classifier (a fully-connected layer followed by softmax). We train this classifier on the global average pooling features of a ResNet. We report top-1 classification accuracy.

\subsubsection{Implementation Details.}
\label{subsec:implementation_details}
We use ResNet-18 and ResNet-50~\cite{7780459} as our encoder to be consistent with the existing literature~\cite{He_2020_CVPR,chen2020simple}. Especially, for CIFAR-100, we use ResNet*. We train models for 300 epochs on CIFAR-100 and for 200 epochs on Tiny ImageNet. We conducted the training on a single Nvidia V100 GPU. A more detailed explanation can be found in the supplementary materials. 
\subsubsection{Results on PNDA for Contrastive Learning.}
\label{subsubsec:results_on_pnda}
\begin{table*}[t]
 \caption{Top-1 linear classification accuracies of rotation PDA, NDA, PNDA for MoCo v2 and SimCLR on CIFAR-100. The scores are averaged over 3 trials. RP denotes the ratio of positive rotated images. These results show that PDA and NDA might degrade the performance, but rotation PNDA boosts the performance of contrastive learning.}
  \centering
  \scalebox{1.1}{
  \begin{tabular}{ll|lll|l}
    \hline
       & & \multicolumn{1}{c}{\;\;\;None \;} & \multicolumn{1}{c}{\;\;\;+ PDA\;\;\;\;\;}  & \multicolumn{1}{c}{\;\;\;+NDA \;\;\;} & \multicolumn{1}{c}{+ PNDA (ours)}\\
      
     & RP (\%) & \multicolumn{1}{c}{-} & \multicolumn{1}{c}{100} & \multicolumn{1}{c}{0} & \multicolumn{1}{c}{12}\\
    \hline\hline  
    MoCo v2~\cite{chen2020mocov2} & ResNet-18* & 62.74$\pm$0.37 &57.18$\pm$0.27\color{red}$\downarrow$5.56\color{black} & 62.75$\pm$0.29&\textbf{63.18$\pm$0.22}\color{blue}$\uparrow$0.44\color{black}\\
    
      & ResNet-50* & 67.51$\pm$0.08 & 63.36$\pm$0.12\color{red}$\downarrow$4.15  \color{black} & 67.28$\pm$0.32 &\textbf{68.20$\pm$0.23}\color{blue}$\uparrow$0.69\color{black}\\
    \hline 
    SimCLR~\cite{chen2020simple} & ResNet-18* & 62.71$\pm$0.38 &61.12$\pm$0.18\color{red}$\downarrow$1.59\color{black} &61.73$\pm$0.23 &\textbf{63.42$\pm$0.04}\color{blue}$\uparrow$0.71\color{black}\\
    
     & ResNet-50* & 65.90$\pm$0.17 &  64.46$\pm$0.09\color{red}$\downarrow$1.44  \color{black} & 64.67$\pm$0.01 &\textbf{66.55$\pm$0.12}\color{blue}$\uparrow$0.65\color{black}\\
     \hline 
  \end{tabular}
  }
  \label{table:results_cifar}
\end{table*}

\begin{table*}[t]
 \caption{Top-1 linear classification accuracies of rotation PDA, NDA, PNDA for MoCo v2 and SimCLR on Tiny ImageNet. The scores are averaged over 3 trials. RP denotes the ratio of positive rotated images. These results show that PDA and NDA might degrade the performance, but rotation PNDA boosts the performance of contrastive learning.}
  \centering
  \scalebox{1.1}{
  \begin{tabular}{ll|lll|l}
    \hline
       & & \multicolumn{1}{c}{\;\;\;None \;} & \multicolumn{1}{c}{\;\;\;+ PDA\;\;\;\;\;}  & \multicolumn{1}{c}{\;\;\;+NDA \;\;\;} & \multicolumn{1}{c}{+ PNDA (ours)}\\
      
     & RP (\%) & \multicolumn{1}{c}{-} & \multicolumn{1}{c}{100} & \multicolumn{1}{c}{0} & \multicolumn{1}{c}{31}\\
    \hline\hline  
    MoCo v2~\cite{chen2020mocov2} &  ResNet-18 & 34.33$\pm$0.23 &30.76$\pm$0.08\color{red}$\downarrow$3.57\color{black} & 34.60$\pm$0.16 & \textbf{35.78$\pm$0.30}\color{blue}$\uparrow$1.45\color{black}\\
    & ResNet-50 & 38.88$\pm$0.40 &35.06$\pm$0.61\color{red}$\downarrow$3.82\color{black} & 38.94$\pm$0.51 & \textbf{39.93$\pm$0.47}\color{blue}$\uparrow$1.05  \color{black}\\
    & ResNet-18* & 45.06$\pm$0.28 &41.42$\pm$0.20\color{red}$\downarrow$3.64 \color{black} & 45.29$\pm$0.20 & \textbf{46.35$\pm$0.10}\color{blue}$\uparrow$1.29  \color{black}\\
    \hline 
    SimCLR~\cite{chen2020simple} & ResNet-18 & 35.91$\pm$0.22 &35.74$\pm$0.18\color{red}$\downarrow$0.17\color{black} & 36.59$\pm$0.14 &\textbf{37.17$\pm$0.15}\color{blue}$\uparrow$1.26\color{black}\\
    & ResNet-50 & 40.10$\pm$0.30 &40.00$\pm$0.20\color{red}$\downarrow$0.10\color{black} & 41.07$\pm$0.13 &\textbf{41.48$\pm$0.24}\color{blue}$\uparrow$1.38\color{black}\\
     \hline 
  \end{tabular}
  }
  \label{table:results_tin}
\end{table*}

\begin{figure}[t]
  \centering
  \includegraphics[keepaspectratio, scale= 0.45]
  {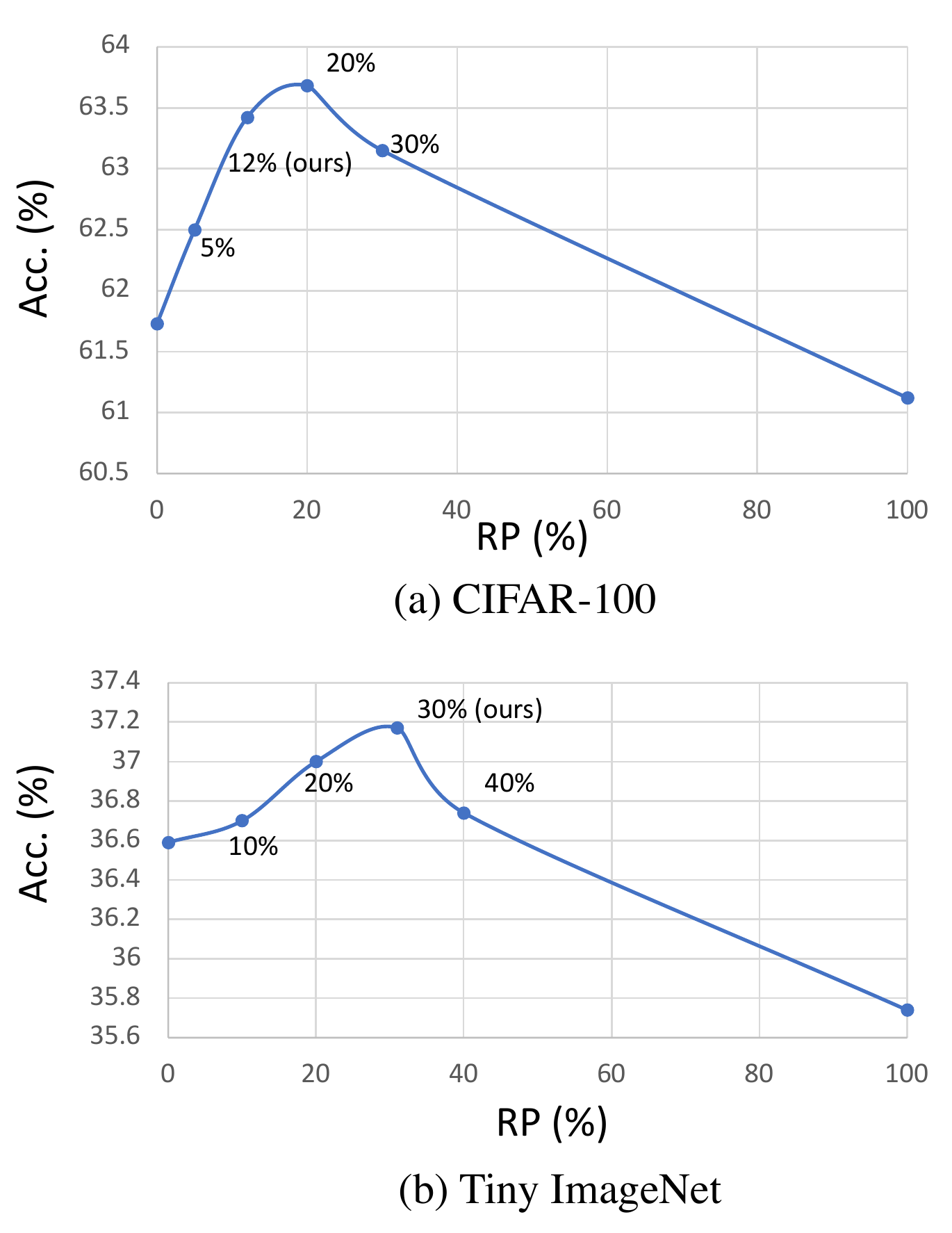}
  \caption{The effect of the ratio of positive rotated images on CIFAR-100 and Tiny ImageNet. The results show that our sampling method can extract approximately the correct number of RAI images.}
  \label{fig:ratio_RP}
\end{figure}

\begin{table*}[t]
\caption{Ablation of each element of PNDA. We use MoCo v2 with ResNet18* on CIFAR-10}
  \centering
  \scalebox{1.1}{
  \begin{tabular}{l|c|c|c}
    \hline
    Methods   & positive for RAI& negative for non-RAI & acc (\%)\\
    \hline \hline
    MoCo v2 & - & - & 62.74\\ 
    \hline 
    +  positive & \checkmark & - & \underline{62.92}\\
    \hline 
    + positive or negative (PNDA) & \checkmark & \checkmark & \textbf{63.18}\\
    \hline 
  \end{tabular}
  }
  \label{table:ablation_pnda}
\end{table*}
\begin{table*}[t]
\caption{Top-1 linear classification accuracies of rotation PDA, NDA, PNDA for BYOL. The scores are averaged over 3 trials on CIFAR-100.}
  \centering
  \scalebox{1.1}{
  \begin{tabular}{ll|ccc|c}
    \hline
     & & \;\;\; None \; & \;\;\;+ PDA\;\;\;\;\;  & \;\;\;+NDA \;\;\; & + PNDA (ours)\\
 
    & RP (\%) & - & 100 & 0 & 12\\
    \hline\hline  
    BYOL~\cite{grill2020bootstrap} & ResNet-18* & 60.81$\pm$0.16 & 57.11$\pm$0.23\color{red}$\downarrow$3.70 \color{black} & 60.51$\pm$0.47 & \textbf{61.68$\pm$0.47} \color{blue}$\uparrow$0.87\color{black}\\
    \hline 
  \end{tabular}
  }
  \label{table:byol_pnda}
\end{table*}

Table~\ref{table:results_cifar}, \ref{table:results_tin} show the results of rotation PDA, NDA and PNDA for MoCo v2 and SimCLR.  We found that rotation PDA degrades the performance in all experiments. Rotation NDA outperforms the baselines of MoCo v2 and SimCLR in some settings, but the differences between them are not
large. However, our proposal PNDA outperforms all com-
parison methods in all experiments, although PNDA only treats rotation as positive for a few images (12\% for CIFAR-
100 and 31\% for Tiny ImageNet) and negative for the other

images.
\subsection{Ablation Studies}
\textbf{The ratio of positive rotated images.}
We examined the effect of the ratio of positive rotated images. 0, 5, 20, 30, and 100\% of the images on CIFAR-100 and 0, 10, 20, 40, and 100\% of the images on Tiny ImageNet in the descending order of the score are treated as positive rotated images. Then, we use RAI extracted by our sampling framework (12\% for CIFAR-100 and 31\% for Tiny ImageNet) and compare the accuracies. Fig.~\ref{fig:ratio_RP} shows the results of our experiments with SimCLR using ResNet18*. The experimental results show that the number of RAI extracted by our sampling framework is close to optimal. This result also demonstrates the validity of tuning the hyperparameters of our sampling method in Section~\ref{subsec:criterion_for_tune}.

\textbf{The effect of each element of PNDA.}
We investigated the effectiveness of each element of PNDA. Table~\ref{table:ablation_pnda} shows the comparison results for MoCo v2 with ResNet18* on CIFAR-100. The results show that both the processes of treating RAI's rotated images as positives and non-RAI's rotated images as negatives contribute to the high performance of PNDA. This result indicates the necessity of processing each image separately for RAI and non-RAI.

\subsection{PNDA for BYOL}
Our PNDA can be applied to contrastive learning frameworks without negatives such as BYOL~\cite{grill2020bootstrap}. Methods, such as BYOL~\cite{grill2020bootstrap}, do not rely on negatives. BYOL minimizes their negative cosine similarity between positives. With the embedding feature $z_i$ and $z_p$ in Section~\ref{subsec:contrastive_lr}, the loss for BYOL is defined as follows: 
\begin{equation}
{\mathcal{L}_{i}}^{\rm{BYOL}}= \|z_i - z_p\|. 
\end{equation}
For BYOL, we refer to \cite{ge2021robust}, which incorporates patch-based NDA into BYOL. We extended \cite{ge2021robust} for rotation PNDA. We define the set of rotated positive samples ${P_i}^{r'}$ and rotated negative samples ${N_i}^{r'}$ which are rotated images of an anchor $x_i$. The extended BYOL loss for PNDA is defined as follows: 
\footnotesize
\begin{equation}
\begin{aligned}
  {\mathcal{L}_{i}}^{\rm{PNDA}} &=\\ \|z_i - z_p\|& + \frac{1}{|{P_i}^{r'}|} {\sum_{{z_{p'}}\in {P_i}^{r'}}}\|z_i- {z_{p'}}\| - \frac{\alpha}{|{N_i}^{r'}|} {\sum_{{z_{n}}\in {N_i}^{r'}}}\|z_i- {z_{n}}\|,
  \label{eq:pnda_for_byol}
  \end{aligned}
\end{equation}
\normalsize

where $\alpha$ is the parameter that controls the penalty on the similarity between the representations of the anchor image and the negative rotated images. We set $\alpha$ to 0.05. We set ${P_i}^{r'}$ as Rot($\hat{{x_i}}^{+}$, 90), Rot($\hat{{x_i}}^{+}$, 180) and Rot($\hat{{x_i}}^{+}$, 270) for RAI and $\phi$, which denotes no images, for non-RAI. We set ${N_i}^{r'}$ as $\phi$ for RAI and Rot($\hat{{x_i}}^{+}$, 90), Rot($\hat{{x_i}}^{+}$, 180) and Rot($\hat{{x_i}}^{+}$, 270) for non-RAI. 

Table~\ref{table:byol_pnda} shows the results for BYOL. We found that our proposal PNDA improves the performance. 

\section{Discussion}
\label{sec:discuttion}
\subsection{Limitations}
\label{subsec:limitation}
The performance of our proposal PNDA depends on the RAI sampling results. In the previous section, we showed that PNDA boosts the performance of contrastive learning. However, the sampling results could be improved. We extracted RAI by focusing on the difficulty of predicting image rotation, but we cannot consider some issues, such as the background dependencies~\cite{xiao2020noise} or the case of multiple objects in an image~\cite{deng2009imagenet}. Large-scale datasets, such as ImageNet~\cite{deng2009imagenet}, have these issues and require more accurate sampling methods.
By developing a more accurate sampling method, the performance of PNDA can still be improved. Addressing such issues is a future challenge.
\subsection{Extentions}
\label{subsec:extantions}
To the best of our knowledge, this work is the first attempt to determine whether an image is rotation-invariant or rotation-variant. Our method can be generalized to many rotation-based methods, not limited to contrastive learning. Furthermore, in this work, we focus on rotation. In addition, the problem of augment-invariance exists in various augmentations other than rotation. Therefore, it is intriguing to consider generalizing our PNDA to apply to other augmentations. 

\section{Conclusion}
In this paper, we propose a novel augmentation strategy called adaptive Positive or Negative Data Augmentation (PNDA), which treats rotation as the better one of either positive or negative considering the semantics of each image. To achieve PNDA, we tackle a novel task for sampling rotation-agnostic images for which rotation is treated as positive. Our experiments demonstrated that rotation PNDA improves the performance of contrastive learning.
PNDA might increase accuracy in augmentation other than rotation, which was previously considered ineffective. We think this perspective will facilitate future work.
\subsection*{Acknowledgement}
This work was partially supported by JST JPMJCR22U4 and JSPS KAKENHI 20J22372, Japan.

\section{ Experimental details}
Detailed experimental setups of SimCLR, MoCo v2 and BYOL are given below.
\begin{enumerate}
   \item \textbf{SimCLR.} We use ResNet-18* and ResNet-50* for CIFAR-100 and use ResNet-18 and ResNet-50 for Tiny ImageNet. These networks are followed by the two-layer multilayer perceptron (MLP) projection head (output dimensions are 128).
   We set $\tau$ to 0.5 in all experiments. For data augmentations, we adopt basic augmentations proposed by Chen et al.~\cite{chen2020simple}: namely, inception crop, horizontal flip, color jitter, and grayscale. 
   On CIFAR-100, models are trained for up to 300 epochs with a batch size of 128. On Tiny ImageNet, models are trained for 200 epochs with a batch size of 256. 
   For optimization, we train under LARS optimizer~\cite{you2017large} with a weight decay of 1e-6 and a momentum with 0.9. An initial learning rate is 0.20.
   For the learning rate scheduling, we use linear warmup~\cite{goyal2017accurate} for early 10 epochs and decay with cosine decay schedule without a restart~\cite{loshchilov2016sgdr}.
   For evaluation, we train a linear classifier for 90 epochs with a batch size of 128 using stochastic gradient descent with a momentum of 0.9 in all experiments. The learning rate starts at 0.1 and is dropped by a factor of 10 at 60\%, 75\%, and 90\% of the training progress. We conducted the training on a single Nvidia V100 GPU.
   \item \textbf{MoCo v2.} We use ResNet-18* and ResNet-50* for CIFAR-100 and ResNet-18, ResNet-50 and ResNet-18* for Tiny ImageNet. 
   These networks are followed by the two-layer multilayer perceptron (MLP) projection head (output dimensions are 128). We set $\tau$ to 0.2 in all experiments.  For data augmentations, we adopt SimCLR's basic augmentations for CIFAR-100. For Tiny ImageNet, we use MoCo v2's augmentation (to add gaussian blur to basic augmentations). The memory bank size is 4096. The momentum for the exponential moving average (EMA) update is 0.999. On CIFAR-100, models are trained for up to 300 epochs with a batch size of 128. On Tiny ImageNet, models are trained for 200 epochs with a batch size of 256. 
   For optimization, we use stochastic gradient descent with a momentum of 0.9 and a weight decay of 1e-4 in all experiments. An initial learning rate is 0.125.
   For the learning rate scheduling, we use cosine decay schedule.
   For evaluation, we train a linear classifier for 90 epochs with a batch size of 128 in all experiments. An initial learning rate for linear evaluation is chosen among \{0.5, 1.5, 2.5, 5, 15, 25, 35\} and is dropped by a factor of 10 at 60\%, 75\%, and 90\% of the training progress. We conducted the training on four Nvidia V100 GPUs.
\item \textbf{BYOL.}
    We use ResNet-18*. These networks are followed by the two-layer multilayer perceptron (MLP) projection head (output dimensions are 128). The momentum for the exponential moving average (EMA) update is 0.999. We do not symmetrize the BYOL loss. For data augmentations, we adopt SimCLR's basic augmentations. Models are trained for up to 300 epochs with a batch size of 128.
    For optimization, we use stochastic gradient descent with a momentum of 0.9 and a weight decay of 1e-4 in all experiments. An initial learning rate is 0.125.
    For the learning rate scheduling, we use cosine decay schedule.
    For evaluation, we train a linear classifier for 90 epochs with a batch size of 128 in all experiments. An initial learning rate for linear evaluation is chosen among \{0.5, 1.5, 2.5, 5, 15, 25, 35\} and is dropped by a factor of 10 at 60\%, 75\%, and 90\% of the training progress. We conducted the training on four Nvidia V100 GPUs.
\end{enumerate}
The setup of PDA, NDA and PNDA for SimCLR, MoCo v2 and BYOL are the same as above, respectively. 

\vskip\baselineskip

{\small
\bibliographystyle{ieee_fullname}
\bibliography{egbib}
}
\end{document}